\documentclass[11pt]{article}

\usepackage[letterpaper, margin=1in, top=1in, bottom=1in]{geometry}
\usepackage{times}
\usepackage{titlesec}
\titleformat{\section}{\large\bfseries}{\thesection}{0.8em}{}
\titleformat{\subsection}{\normalsize\bfseries}{\thesubsection}{0.8em}{}
\titleformat{\subsubsection}{\normalsize\itshape}{\thesubsubsection}{0.8em}{}
\titlespacing{\section}{0pt}{10pt}{4pt}
\titlespacing{\subsection}{0pt}{8pt}{3pt}

\usepackage{fancyhdr}
\fancypagestyle{plain}{\fancyhf{}\fancyfoot[C]{\small\thepage}}

\usepackage[utf8]{inputenc}
\usepackage[T1]{fontenc}
\usepackage[backref]{hyperref}
\usepackage{url}
\usepackage{booktabs}
\usepackage{amsfonts}
\usepackage{nicefrac}
\usepackage{microtype}
\usepackage[table]{xcolor}
\definecolor{lightgraybg}{gray}{0.95}
\definecolor{darkgraybg}{gray}{0.78}

\usepackage{graphicx}
\usepackage{subcaption}
\usepackage{enumitem}
\usepackage{multirow}
\usepackage{amsmath}
\usepackage{amssymb}
\usepackage{mathtools}
\usepackage{amsthm}
\usepackage{algorithm}
\usepackage{algorithmic}
\usepackage{wrapfig}
\usepackage[capitalize,noabbrev]{cleveref}
\usepackage[textsize=tiny]{todonotes}
\usepackage[most]{tcolorbox}
\usepackage{alltt}
\usepackage[export]{adjustbox}
\usepackage{float}

\hypersetup{hidelinks, colorlinks=true, linkcolor=brown, citecolor=brown}

\newtcolorbox{AssistantBox}[2][]{assistantbox,title=#2,#1}
\newtcolorbox{UserBox}[2][]{userbox,title=#2,#1}
\newtcolorbox{AIBox}[2][]{aibox,title=#2,#1}
\newtcolorbox{AIBoxBreak}[2][]{aiboxbreakable,title=#2,#1}

\tcbset{
  assistantbox/.style={width=\linewidth,top=10pt,colback=black!05,colframe=black!40,
    colbacktitle=black!50,enhanced,center,
    attach boxed title to top left={yshift=-0.1in,xshift=0.15in},
    boxed title style={boxrule=0pt,colframe=white}}}
\tcbset{
  aibox/.style={width=\linewidth,top=10pt,colback=black!05,colframe=black!20,
    colbacktitle=black!50,enhanced,center,
    attach boxed title to top left={yshift=-0.1in,xshift=0.15in},
    boxed title style={boxrule=0pt,colframe=white}}}
\tcbset{
  aiboxbreakable/.style={width=\linewidth,top=10pt,colback=black!05,colframe=black!20,
    colbacktitle=black!50,enhanced,center,breakable,
    attach boxed title to top left={yshift=-0.1in,xshift=0.15in},
    boxed title style={boxrule=0pt,colframe=white}}}

\theoremstyle{plain}

\theoremstyle{definition}

\theoremstyle{remark}

\title{Scalable Token-Level Hallucination Detection in Large Language Models}

\author{
  Rui Min$^{1,2}$ \quad Tianyu Pang$^{1}$ \quad Chao Du$^{1}$ \quad Minhao Cheng$^{3}$ \quad Yi R. Fung$^{2}$ \\[4pt]
  {\small $^1$Sea AI Lab \quad $^2$Hong Kong University of Science and Technology \quad $^3$Pennsylvania State University}
}

\date{}

\begin{document}

\maketitle

\begin{abstract}
Large language models (LLMs) have demonstrated remarkable capabilities, but they still frequently produce hallucinations. These hallucinations are difficult to detect in reasoning-intensive tasks, where the content appears coherent but contains errors like logical flaws and unreliable intermediate results. While step-level analysis is commonly used to detect internal hallucinations, it suffers from limited granularity and poor scalability due to its reliance on step segmentation. To address these limitations, we propose \textbf{\textsc{TokenHD}}, a holistic pipeline for training token-level hallucination detectors. Specifically, \textsc{TokenHD} consists of a scalable data engine for synthesizing large-scale hallucination annotations along with a training recipe featuring an importance-weighted strategy for robust model training. To systematically assess the detection performance, we also provide a rigorous evaluation protocol. Through training within \textsc{TokenHD}, our detector operates directly on free-form text to identify hallucinations, eliminating the need for predefined step segmentation or additional text reformatting. Our experiments show that even a small detector (0.6B) achieves substantial performance gains after training, surpassing much larger reasoning models (e.g., QwQ-32B), and detection performance scales consistently with model size from 0.6B to 8B. Finally, we show that our detector can generalize well across diverse practical scenarios and explore strategies to further enhance its cross-domain generalization capability.\footnote{Code is available at \url{https://github.com/rmin2000/TokenHD}.}
\end{abstract}

\section{Introduction}

Large language models (LLMs) have achieved remarkable performance on reasoning-intensive tasks, such as solving complex mathematical problems~\cite{shao2024deepseekmath, guo2025deepseek,comanici2025gemini} and code generation~\cite{hui2024qwen2, anthropic_claude_sonnet_4_5}. Nevertheless, LLMs still suffer from hallucinations~\cite{ji2023survey, huang2025survey, kalai2025language}: generated content may appear coherent yet contain factual errors or logical inconsistencies, making such errors difficult to detect and undermining response reliability.

To address these challenges, various post-hoc detection methods have been proposed to scrutinize the truthfulness of LLM-generated content. Earlier research primarily focused on factual hallucinations~\cite{chen2023hallucination, manakul2023selfcheckgpt, obeso2025real}, such as detecting contradictions between generated statements and trusted knowledge sources. However, with the emergence of retrieval-augmented generation (RAG) systems~\cite{gao2023retrieval} and the integration of search tools~\cite{li2025search}, these knowledge-based hallucinations can be largely mitigated. In contrast, hallucinations in reasoning-intensive tasks (e.g., mathematics) often manifest as subtle logical errors or incorrect intermediate results, making them significantly harder to detect. To locate these errors, a common practice is to decompose a solution into individual steps and train a detector with step-level supervision. Specifically, Process Reward Models (PRMs)~\cite{lightman2023let, wang2024math, zhang2025lessons} adopt this paradigm by assigning a correctness label to each step, thereby pinpointing where the internal reasoning process goes wrong.

\begin{wrapfigure}{r}{0.55\textwidth}
    \centering
    \includegraphics[width=0.44\textwidth]{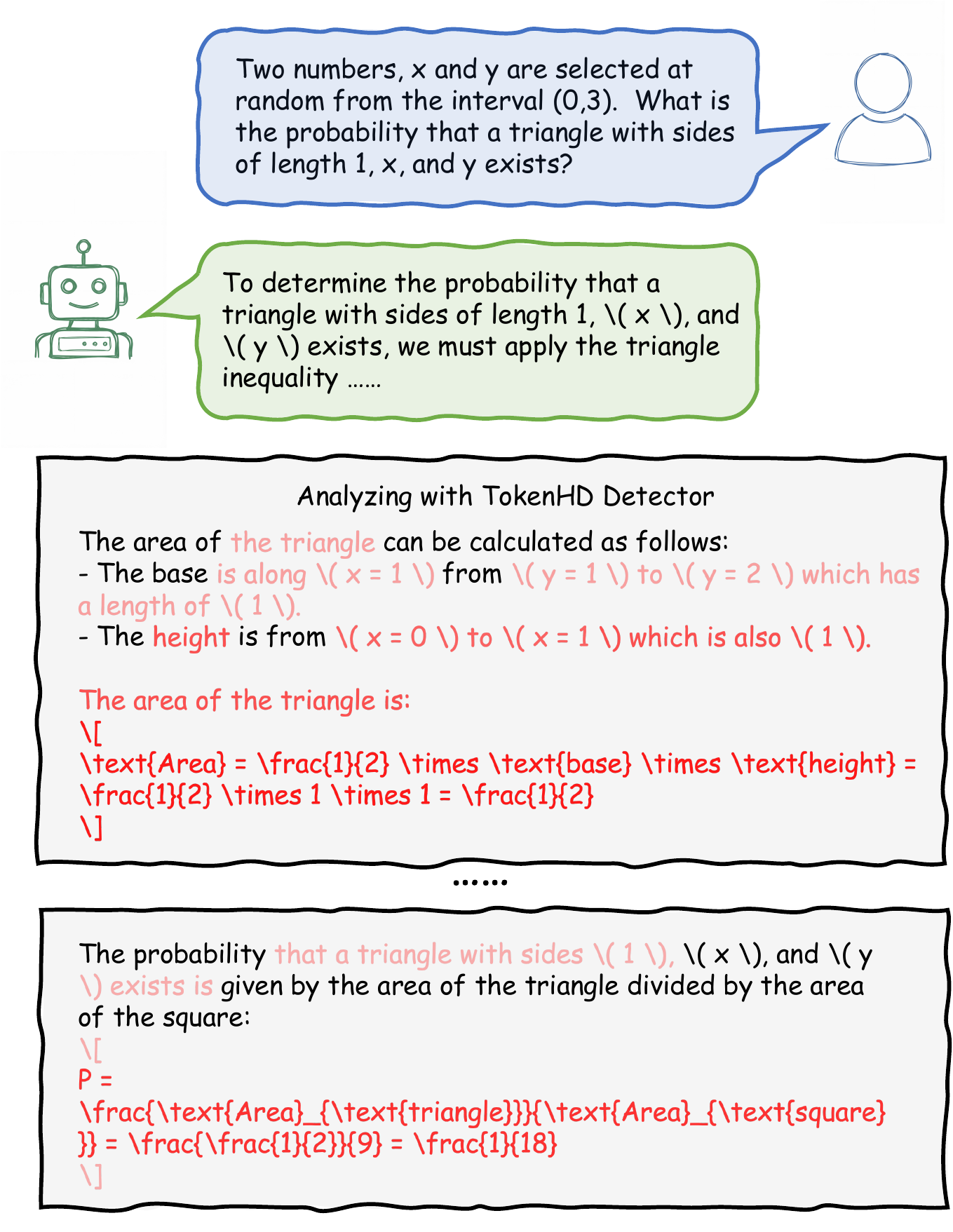}
    \caption{An illustration of the token-level detection mechanism of \textbf{\textsc{TokenHD}}. Our detector identifies hallucinations directly in free-form text without predefined step segmentation. Color intensity reflects predicted hallucination probability: \textcolor{red!80!black}{\textbf{deeper highlights}} indicate higher likelihood, \textcolor{red!30}{\textbf{lighter highlights}} lower likelihood.}
    \label{fig:demo_figure}
\end{wrapfigure}

Nevertheless, PRMs face several limitations because they necessitate explicit step segmentation, which is difficult when model outputs are free-form or lack clear separation boundaries. Furthermore, they are inherently restricted to step-level analysis, lacking fine-grained and flexible hallucination localization. While search-based methods such as Monte Carlo Tree Search (MCTS)~\cite{wang2024math, xie2024monte, zhang2025lessons} can estimate intermediate correctness via sampling statistics, they incur prohibitive computational overhead from intensive policy-model queries, limiting scalability. To address these challenges, we shift our focus to the atomic units of text by proposing \textsc{TokenHD}, a holistic framework for token-level hallucination detection. Unlike PRMs, \textsc{TokenHD} operates directly on free-form text and evaluates hallucination for each token. This design enables the precise location of hallucinations while substantially reducing inference latency, since our detector assigns scores directly without the need to generate verbal analysis. Figure~\ref{fig:demo_figure} shows an example of our detection mechanism in a mathematical task.

The proposed \textsc{TokenHD} comprises a scalable data synthesis engine and a specialized training recipe. To overcome the scarcity of token-level hallucination labels in existing datasets, we first develop a data synthesis pipeline to obtain high-quality samples with token-level annotations. Specifically, for each candidate sample, we prompt multiple critic models to identify hallucinated text fragments. These fragments are refined through a text restoration process, converted into token space, and merged into a single label sequence through averaging. To improve the aggregation quality beyond simple averaging, we further design an adaptive ensemble strategy that optimizes weights for each critic model to perform weighted aggregation. Leveraging these token-level supervisions, we train our detector using an importance-weighted strategy specifically designed to address the sparsity of hallucinated tokens. We also establish a rigorous evaluation protocol to assess detection performance and conduct extensive experiments on mathematical and STEM benchmarks. Finally, we show that our detector can generalize well across diverse task domains and policy models, while also exploring several practical strategies to enhance its generalization capability.

\section{Preliminaries}
\label{sec:preliminaries}
We first introduce the basic concepts of tokenization and detokenization. We then formalize our token-level hallucination detection task, and finally describe the evaluation metrics used to assess detection performance.

\textbf{Tokenization and Detokenization.} Given an input query $\mathbf{x}$, a policy model $\pi(\cdot)$ generates an output $\mathbf{y} = \pi(\mathbf{x})$. The tokenization function $\tau(\cdot)$ maps $\mathbf{y}$ to a token sequence $\mathbf{t}=\tau(\mathbf{y})=(t_i)_{i=1}^{|\tau(\mathbf y)|}$ with length $|\tau(\mathbf y)|$, where $t_i\in\{0,1,\dots,|\mathcal{V}|-1\}$ and $\mathcal{V}$ is the vocabulary. Conversely, the detokenization function $\tau^{-1}(\cdot)$ maps each token to a string fragment $\tau^{-1}(t_i)$, and the reconstructed text is obtained by string concatenation: $\mathbf{y}=\texttt{concat}_{i=1}^{|\tau(\mathbf y)|} \tau^{-1}(t_i)$.

\textbf{Token-Level Hallucination Detection.} We define token-level hallucination detection as predicting a hallucination score for each token position in $\mathbf{t}$. Let $\mathcal{H}(\cdot)$ denote a hallucination detector; it outputs a score sequence $\widehat{\mathbf{s}}=\mathcal{H}(\mathbf{x},\mathbf{y})=(\widehat{s}_i)_{i=1}^{|\tau(\mathbf y)|}$, where $\widehat{s}_i\in[0,1]$ and larger values indicate a higher likelihood of hallucination. Since $\widehat{s}_i$ is a continuous score, to map the predictions back to text fragments, we introduce a threshold $\beta_{\widehat{I}}$ and define the predicted hallucinated token indexes as $\widehat{I}=\{i\mid \widehat{s}_i>\beta_{\widehat{I}}\}$. We then group consecutive indices in $\widehat{I}$ into $M$ segments $\{\widehat{I}_m\}_{m=1}^{M}$, and map each segment back to its corresponding text span $\widehat{r}_m=\texttt{concat}_{i\in \widehat{I}_m}\tau^{-1}(t_i)$.

\textbf{Evaluation Metrics.} To evaluate our detector, we compare the predicted token-level scores $\widehat{\mathbf{s}}$ against the ground-truth annotations $\mathbf{s}=(s_i)_{i=1}^{|\tau(\mathbf y)|}$. The ground-truth $\mathbf{s}$ is provided by labeler models, typically high-capacity models (e.g., GPT-5). Due to their expensive costs, these models are solely used for annotating small-scale evaluation samples. For ease of evaluation, we binarize the ground-truth scores with a threshold $\beta_I$, yielding ground-truth hallucinated token indexes $I=\{i \mid s_i>\beta_I\}$. We then compute token-level precision, recall, and $\mathrm{F}_1$ as
$\mathrm{Precision}=\frac{|\widehat{I}\cap I|}{|\widehat{I}|}$,
$\mathrm{Recall}=\frac{|\widehat{I}\cap I|}{|I|}$, and $\mathrm{F}_1=\frac{2\,\mathrm{Precision}\cdot \mathrm{Recall}}{\mathrm{Precision}+\mathrm{Recall}}$. These token-level metrics provide a fine-grained measure of detection performance and serve as our primary evaluation metrics.

\vspace{-0.15cm}
\section{The \textsc{TokenHD} Framework}
\vspace{-0.15cm}
We present the core idea of our \textsc{TokenHD} framework, including how we obtain samples with token-level hallucination annotations and the techniques used for training detectors. In Section~\ref{subsec:token_annotation}, we describe how we obtain our token-level hallucination annotations to construct the dataset. Then, in Section~\ref{subsec:token_ensemble}, we introduce an ensemble strategy that adaptively and optimally aggregates diverse sources of annotations to improve the annotation quality. Finally, we detail how we leverage the annotations to train the token-level hallucination detector in Section~\ref{subsec:detector_training}.

\subsection{Token‑Level Hallucination Annotations}
\label{subsec:token_annotation}
We start by introducing our data construction process. While labeler models offer high-quality annotations, their prohibitive inference costs make them impractical for large-scale data generation. Therefore, we instead employ more cost-effective critic models $\mathcal{V}(\cdot)$ to produce our training data. Given an input query $\mathbf{x}$ and the corresponding output $\mathbf{y}=\pi(\mathbf{x})$, we prompt a critic model $\mathcal{V}(\cdot)$ to identify hallucinated text fragments in $\mathbf{y}$. The model returns a sequence of $M$ hallucinated text fragments $\mathcal{V}(\mathbf{x}, \mathbf{y})=( r_m^{\mathcal{V}} )_{m=1}^{M}$. Note that $M=0$ if no hallucinated content is identified by the critic model, which corresponds to an all-zero annotation sequence. In practice, raw annotations may be paraphrased or slightly expanded, making direct token-level alignment difficult. To address this, we design a restoration process. We prompt an LLM to iteratively refine the raw fragments so that each hallucinated fragment corresponds to an exact span in the original $\mathbf{y}$. Experiments demonstrate that our strategy achieves 98.10\% recovery performance on average across critics. We provide details of the annotation process and our restoration algorithm in Appendix~\ref{appendix:data_annotation}. After restoration, we convert the restored fragment-level annotations into token-level annotations over the token sequence $\tau(\mathbf{y})=(t_i)_{i=1}^{|\tau(\mathbf y)|}$. Specifically, we map each token $t_i$ to its detokenized string fragment $\tau^{-1}(t_i)$ in $\mathbf{y}$, and assign a binary hallucination label $a_i^{\mathcal{V}}\in\{0,1\}$. We denote $\mathcal{A}_{\mathcal{V}}$ as the hallucination annotations produced by critic model $\mathcal{V}$:

\begin{equation}
    \begin{aligned}
    \mathcal{A}_{\mathcal{V}} &= \bigl\{(t_i, a_i^{\mathcal{V}})\bigr\}_{i=1}^{|\tau(\mathbf y)|}\text{,} \\
    a_i^{\mathcal{V}} &=
    \begin{cases}
    1\text{,} & \text{if } \tau^{-1}(t_i) \text{ overlaps with } r_m^{\mathcal{V}} \text{ for some } m\text{,} \\[4pt]
    0\text{,} & \text{otherwise.}
    \end{cases}
    \end{aligned}
    \label{eq:hallu_ann}
\end{equation}
Here, ``overlaps'' means that the text spans of \(\tau^{-1}(t_i)\) and \(r_m^{\mathcal V}\) in \(\mathbf y\) have a non-empty intersection. And when there are no hallucinations within $\mathbf{y}$, i.e., $M=0$, the annotations reduce to $a_i^{\mathcal{V}}=0$ for all $i$. To obtain more reliable token-level hallucination labels and mitigate the variance from a single critique, we perform $C$ critiques for each output $\mathbf{y}$ using the same critic model $\mathcal{V}$. For each critique $c \in \{1, \dots, C\}$, we obtain the corresponding token-level binary annotations $\mathcal{A}_{\mathcal{V}}^{(c)} = \{(t_i, a_{i}^{\mathcal{V},(c)})\}_{i=1}^{|\tau(\mathbf y)|}$ as defined in Eq.~\ref{eq:hallu_ann}. We then aggregate these $C$ binary annotations into soft hallucination scores by averaging across critiques, denoted as $\bar{a}_i^{\mathcal{V}} = \frac{1}{C} \sum_{c=1}^{C} a_{i}^{\mathcal{V},(c)}$ and obtain the annotations $\bar{\mathcal{A}}_{\mathcal{V}} = \{(t_i,\bar{a}_i^{\mathcal{V}})\}_{i=1}^{|\tau(\mathbf y)|}$.

\subsection{Ensemble from Diverse Annotations}
\label{subsec:token_ensemble}
While performing multiple critiques helps reduce variance, relying solely on a single critic model may still make the annotations strongly affected by the model's capability and preference. In practice, we can leverage multiple models to detect hallucinations, reducing the risk that the bias of any critic model overly influences the annotations. Given a set of $K$ diverse critic models $\{\mathcal{V}_k\}_{k=1}^K$, the hallucination annotations derived from each critic model are $\bar{\mathcal{A}}_{\mathcal{V}_k}=\{(t_i, \bar{a}_i^{\mathcal{V}_k})\}_{i=1}^{|\tau(\mathbf y)|}$. We therefore need to aggregate these annotations into a single label sequence. To address this, we consider two ensemble strategies:

\textbf{Uniform Ensemble.}
A straightforward strategy is to average the hallucination scores across all $K$ critic models for each token, which is defined as: $\bar{\mathcal{A}}_{\text{avg}} = \left\{\left(t_i, \frac{1}{K} \sum_{k=1}^{K} \bar{a}_i^{\mathcal{V}_k}\right)\right\}_{i=1}^{|\tau(\mathbf y)|}$.
This simple scheme directly averages the annotations from different critic models by treating them equally. However, these models can differ in their annotation capability, and directly averaging their hallucination scores may result in a ``barrel effect'', where weaker critic models disproportionately degrade the quality of the averaged annotations. This motivates us to further consider an adaptive ensemble mechanism that accounts for the different annotation capabilities of individual critic models.

\textbf{Adaptive Ensemble.}
We introduce an adaptive ensemble strategy that assigns a learnable weight to each critic model. Let $\mathbf{w} = (w_1,\dots,w_K)$ denote the weight vector, where $w_k$ indicates the contribution of $\mathcal{V}_k$. We learn $\mathbf{w}$ on a validation set $\mathcal{D}_{\mathrm{val}}$. For each validation sample $(\mathbf{x},\mathbf{y},\mathbf{s})\in\mathcal{D}_{\mathrm{val}}$, where $\mathbf{s}$ indicates ground-truth hallucination scores, we first leverage the $K$ critic models to obtain their annotation scores $\{(t_i,\bar{a}_{i}^{\mathcal{V}_k})\}_{i=1}^{|\mathbf{t}|}$ respectively. We then optimize $\mathbf{w}$ by minimizing the Mean Squared Error (MSE) between the ensembled annotation scores and the ground-truth scores:
\begin{equation}
\mathcal{L}_{\mathrm{e}}(\mathbf w) = \mathbb{E}_{(\mathbf x,\mathbf y,\mathbf s)\sim \mathcal{D}_{\mathrm{val}}} \left[ \frac{1}{|\tau(\mathbf y)|} \sum_{i=1}^{|\tau(\mathbf y)|} \left( s_i - \sum_{k=1}^{K} w_k\, \bar{a}_i^{\mathcal V_k} \right)^2 \right]\text{, \quad s.t.} \quad \sum_{k=1}^{K} w_k = 1 \text{.}
\label{eq:ensemble_loss}
\end{equation}
Using the optimized weights $\mathbf{w}^{*} = \arg\min_{\mathbf{w}} \mathcal{L}_{\text{e}}(\mathbf{w})$, we compute the ensembled hallucination score for each token $t_i$ as $\sum_{k=1}^{K} w^{*}_k \bar{a}_i^{\mathcal{V}_k}$, and obtain the adaptively ensembled annotation scores as: $\bar{\mathcal{A}}_{\text{adapt}}
    = \left\{\left(t_i, \sum_{k=1}^{K} w^{*}_k \,\bar{a}_i^{\mathcal{V}_k}\right)\right\}_{i=1}^{|\tau(\mathbf y)|}$.

Compared with the uniform ensemble, this adaptive scheme assigns smaller weights to critic models whose annotations are less consistent with the ground-truth scores on the validation set, thereby reducing the influence of weaker models and making the final annotations more robust. For simplicity, we define $\bar{\mathbf{a}}^{\mathrm{ens}}=(\bar{a}_i^{\mathrm{ens}})_{i=1}^{|\tau(\mathbf y)|}$ as the ensembled hallucination score under the chosen strategy.

\subsection{Training Recipe for the Hallucination Detector}
\label{subsec:detector_training}

After ensembling annotation scores from diverse critic models, we construct training examples and train the hallucination detector $\mathcal{H}(\cdot)$ on the training set $\mathcal{D}_{\mathrm{train}}$. For each token position $i$, we use the ensembled score $\bar{a}_i^{\mathrm{ens}}$ as supervision and introduce two training schemes.

\textbf{Standard Training.}
We first consider a straightforward strategy to optimize the detector using standard cross-entropy loss. For each training example with input query $\mathbf{x}$ and output $\mathbf{y}$, the detector outputs $\widehat{\mathbf{s}}=\mathcal{H}(\mathbf{x},\mathbf{y})=(\widehat{s}_i)_{i=1}^{|\tau(\mathbf y)|}$, and we minimize
$L = \mathbb{E}_{(\mathbf x,\mathbf y,\bar{\mathbf{a}}^{\mathrm{ens}})\sim \mathcal{D}_{\mathrm{train}}}
\left[\frac{1}{|\tau(\mathbf y)|} \sum_{i=1}^{|\tau(\mathbf y)|} \ell_s\bigl(\widehat{s}_i, \bar{a}_i^{\mathrm{ens}}\bigr)\right]$,
where $\ell_s\bigl(\widehat{s}_i, \bar{a}_i^{\mathrm{ens}}\bigr) = - \bar{a}_i^{\mathrm{ens}} \log \widehat{s}_i - (1 - \bar{a}_i^{\mathrm{ens}})\log(1 - \widehat{s}_i)$. This training objective allows the detector to learn directly from the soft scores.

\textbf{Importance-weighted Training.}
However, after curating the training examples, we find that hallucinated tokens with high hallucination scores are sparse across the dataset, leading to label imbalance and biasing the detector toward predicting lower hallucination scores. To address this, we propose a reweighting scheme to our original training objective. Let \texttt{pos\_weight} be the proportion of tokens with hallucination scores less than or equal to a threshold $\beta$, and let \texttt{neg\_weight} be the proportion of tokens with scores greater than $\beta$, i.e., $\texttt{pos\_weight} = \frac{\bigl|\{\,i \mid \bar{a}_i^{\mathrm{ens}} \le \beta\,\}\bigr|}{|\tau(\mathbf y)|}$ and $\texttt{neg\_weight} = \frac{\bigl|\{\,i \mid \bar{a}_i^{\mathrm{ens}} > \beta\,\}\bigr|}{|\tau(\mathbf y)|}$. We then define the following weighted token-level cross-entropy loss:

\begin{equation}
\ell_i(\widehat{s}_i, \bar{a}_i^{\mathrm{ens}}) = - \Bigl[ \texttt{pos\_weight}\cdot \bar{a}_i^{\mathrm{ens}}\log \widehat{s}_i + \texttt{neg\_weight}\cdot (1-\bar{a}_i^{\mathrm{ens}})\log(1-\widehat{s}_i) \Bigr]\text{.}
\label{eq:weighted_loss}
\end{equation}
This strategy increases the contribution of hallucinated tokens during training, which effectively mitigates the challenge of label sparsity and stabilizes the training process.

\section{Evaluating the Effectiveness of \textsc{TokenHD}}
\subsection{Experimental Settings}
\label{subsec:exp_settings}
\textbf{Backbone Models.} To ensure our detector remains efficient for practical use, we focus on small-scale backbones from the Qwen3 series~\cite{yang2025qwen3}, with sizes ranging from 0.6B to 8B. Since the detector is primarily designed as a post-hoc hallucination detection module for much larger LLM systems, low inference latency and deployment costs are critical. While performance scales with model size across this range, even the smallest 0.6B variant substantially outperforms larger reasoning models such as QwQ-32B, confirming that a lightweight detector is practically competitive.

\textbf{Training Settings.} We initially train our detector on mathematical tasks and explore its transferability to other domains, such as code generation, in later sections. The training data is curated from three primary sources: Math~\cite{hendrycks2021measuring}, and subsets of AceReason-Math~\cite{chen2025acereason} and Big-Math~\cite{albalak2025big}, providing problems with various difficulty levels. To generate the training samples, we use GPT-4o-mini~\cite{hurst2024gpt} as our policy model and sample two responses for each prompt. For hallucination annotation, we use four critic models: DeepSeek-R1-0528-Qwen3-8B (R1-Qwen3-8B)~\cite{guo2025deepseek}, QwQ-32B~\cite{team2025qwq}, GPT-4.1~\cite{achiam2023gpt}, and o4-mini~\cite{o3_o4_mini} to annotate each training sample independently, and then aggregate their outputs into a single token-level label sequence using the adaptive ensemble strategy. To improve data quality, we apply several filtering strategies. For instance, we discard samples with incorrect final answers even if the critic models identify no hallucinated tokens. All of our experiments are conducted on a single node with 8$\times$ NVIDIA A100 GPUs. We defer more training details, such as the training data composition and the choice of hyperparameters to Appendix~\ref{appendix:training_settings}.

\begin{table}[t]
\centering
\caption{Detection performance of various models across mathematical benchmarks. Our \textsc{TokenHD} variants are trained from their corresponding backbone models (e.g., \textsc{TokenHD}-0.6B is trained based on Qwen3-0.6B). We report the average $S_{\textrm{incor}}$ / $S_{\textrm{cor}}$, respectively.}
\label{tab:math_benchmark}
\resizebox{0.9\textwidth}{!}{
\begin{tabular}{llcccc}
\toprule
\textbf{Category} & \textbf{Model Name} & \textbf{Math-500} & \textbf{AIME-2024} & \textbf{AIME-2025} & \textbf{Olym-Math}  \\
\midrule
\multirow{2}{*}{Labeler Models}
    & GPT-5  & 86.68 / 99.68 & 84.01 / 100.0  &  87.89 / 100.0  &  84.94 / 99.59    \\
    & o3 & 82.03 / 99.86  & 80.44 / 100.0 & 85.88 / 100.0 & 79.36 / 99.77  \\
\midrule

\multirow{4}{*}{Critic Models}
    & R1-Qwen3-8B  & 43.95 / 99.80  & 34.51 / 100.0    & 40.14 / 100.0   & 39.93 / 99.75   \\
    & QwQ-32B & 55.05 / 99.78  & 50.59 / 100.0 & 56.18 / 99.31 & 52.53 / 99.61  \\
    & GPT-4.1 & 46.60 / 98.56 & 50.20 / 99.37  & 56.24 / 100.0 & 49.80 / 96.99  \\
    & o4-mini & 71.39 / 99.93 & 67.67 / 100.0 & 76.09 / 100.0 & 70.32 / 99.80  \\
\midrule

\multirow{4}{*}{Backbone Models}
    & Qwen3-0.6B & \cellcolor{lightgraybg}9.86 / 98.94   & \cellcolor{lightgraybg}9.29 / 100.0  & \cellcolor{lightgraybg}6.46 / 95.84  & \cellcolor{lightgraybg}8.42 / 98.37  \\
    & Qwen3-1.7B & \cellcolor{lightgraybg}24.00 / 99.93  & \cellcolor{lightgraybg}18.99 / 100.0  & \cellcolor{lightgraybg}18.76 / 100.0  & \cellcolor{lightgraybg}19.45 / 99.82  \\
    & Qwen3-4B   & \cellcolor{lightgraybg}42.23 / 99.91 & \cellcolor{lightgraybg}36.78 / 100.0 & \cellcolor{lightgraybg}44.75 / 100.0 & \cellcolor{lightgraybg}39.39 / 99.86\\
    & Qwen3-8B   & \cellcolor{lightgraybg}49.45 / 99.84 & \cellcolor{lightgraybg}47.03 / 100.0 & \cellcolor{lightgraybg}48.92 / 99.72 & \cellcolor{lightgraybg}46.02 / 99.76\\
\midrule

\multirow{4}{*}{Detector Models}
    & \textsc{TokenHD}-0.6B & \cellcolor{darkgraybg}63.64 / 94.83 & \cellcolor{darkgraybg}61.41 / 90.12  & \cellcolor{darkgraybg}64.34 / 89.31  & \cellcolor{darkgraybg}64.99 / 93.21  \\
    & \textsc{TokenHD}-1.7B & \cellcolor{darkgraybg}64.39 / 98.06 & \cellcolor{darkgraybg}59.57 / 96.79 & \cellcolor{darkgraybg}67.02 / 96.41 & \cellcolor{darkgraybg}65.63 / 96.68  \\
    & \textsc{TokenHD}-4B   & \cellcolor{darkgraybg}64.61 / 98.10 & \cellcolor{darkgraybg}63.70 / 96.62 & \cellcolor{darkgraybg}67.32 / 92.66 & \cellcolor{darkgraybg}66.95 / 96.54 \\
    & \textsc{TokenHD}-8B   & \cellcolor{darkgraybg}64.55 / 98.71  & \cellcolor{darkgraybg}63.62 / 98.44 & \cellcolor{darkgraybg}68.88 / 95.84 & \cellcolor{darkgraybg}68.45 / 97.61 \\
\bottomrule
\end{tabular}
}
\vspace{-0.05cm}
\end{table}

\begin{figure}[t]
    \centering
    \includegraphics[width=0.98\textwidth]{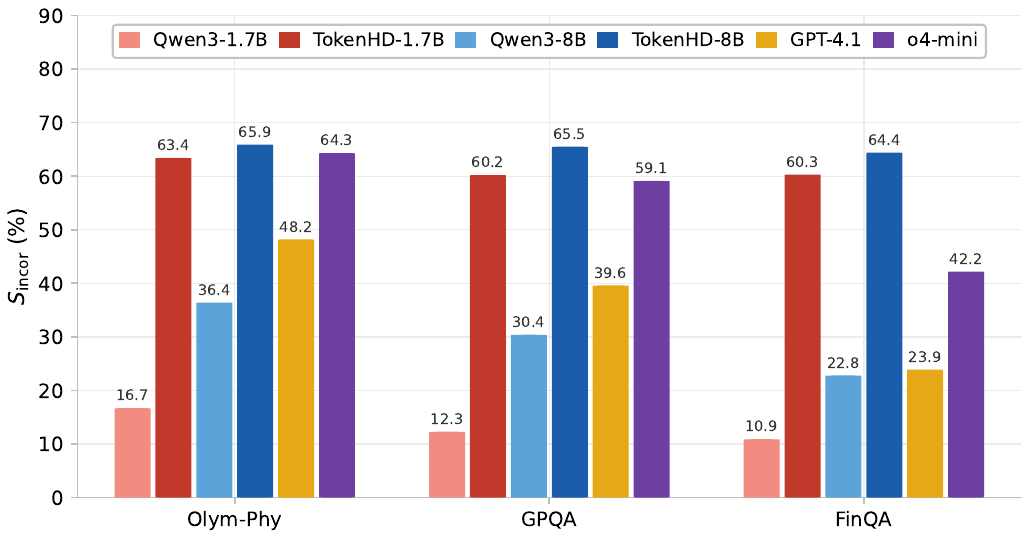}
    \caption{We report $S_{\textrm{incor}}$ across three STEM benchmarks. Qwen3-1.7/8B are backbone models, GPT-4.1 and o4-mini are critic models, and \textsc{TokenHD}-1.7/8B are our trained hallucination detectors.}
    \label{fig:addition_bench_comparison}
    \vspace{-0.25cm}
\end{figure}

\textbf{Evaluation Protocols.} We primarily evaluate our detector on four widely used mathematical benchmarks, including Math-500~\cite{hendrycks2021measuring}, AIME-2024~\cite{aime_2024}, AIME-2025~\cite{aime_2025}, and OlympiadBench-Math~\cite{he2024olympiadbench}, and later extend our evaluation to additional STEM domains, including GPQA~\cite{rein2024gpqa}, OlympiadBench-Phy~\cite{he2024olympiadbench}, and FinQA~\cite{chen2021finqa}. To generate evaluation samples, we use the same policy model as in the training data generation, GPT-4o-mini (we discuss the impact of different policy models in Section~\ref{subsec:policy_task}), and generate two responses for each prompt. We then obtain ground-truth hallucination annotations using two labeler models, GPT-5~\cite{gpt5} and o3~\cite{o3_o4_mini}, and use the uniform ensemble strategy for final aggregation. Following the annotation results, we categorize the evaluation samples into two sets: hallucinated samples, containing tokens identified by the labeler models, and non-hallucinated samples, which are hallucination-free and conclude with a correct answer. We employ token-level F1 (defined in Section~\ref{sec:preliminaries}) as the detection metric for the hallucinated samples, denoted as $S_{\textrm{incor}}$. For non-hallucinated samples, where all ground-truth labels are zero, we first invert the labels to ones and report recall only (because precision is always 1 in this setting), denoted as $S_{\textrm{cor}}$. Both metrics are expressed as percentages, and we report their values as plain numbers throughout this paper. In our experiments, we set $\beta_{I}=0.5$ as the default to generate binary ground-truth labels. For the predicted labels, we use the same threshold $\beta_{\widehat{I}}=0.5$, ensuring a consistent and fair comparison across all models. For comparison, we also report the performance of critic and labeler models by treating their individual annotations as predictions and evaluating them against the aggregated ground-truth labels. We also verify the annotation quality through \textbf{human evaluation} in Appendix~\ref{appendix:localization}, where human raters assigned high-quality scores of 4.63/5 for math and 4.55/5 for code.

\vspace{-0.15cm}
\subsection{Experimental Results}
\textbf{Effectiveness of Token-Level Hallucination Detection.}
As shown in Table~\ref{tab:math_benchmark}, our detectors demonstrate strong performance on the four mathematical benchmarks. We highlight three critical findings from these results. First, our detectors consistently outperform their corresponding backbone models while significantly reducing inference overhead. Unlike critic models, each detector directly outputs token-level scores without generating any reasoning text. Second, although our detector is trained on data labeled by the critic models, it still surpasses most of them, including GPT-4.1, R1-Qwen3-8B, and QwQ-32B, and achieves competitive performance against o4-mini. Notably, our smallest \textsc{TokenHD}-0.6B significantly outperforms QwQ-32B on $S_{\textrm{incor}}$ across all benchmarks. Third, detection performance scales consistently with model size, with 1.7B--8B variants consistently achieving $S_{\textrm{cor}} \geq 92$. These results demonstrate that our training framework enables lightweight models to match much larger reasoning models on this task (see Appendix~\ref{appendix:extended_discussion}). We also provide comparisons against PRM baselines in Appendix~\ref{appendix:prm_comparison}, where \textsc{TokenHD}-8B outperforms Qwen2.5-Math-PRM-72B across five benchmarks. In Appendix~\ref{appendix:auroc}, we report AUROC and AUPRC as metrics, and our results further confirm that \textsc{TokenHD}'s scores provide strong threshold-independent discrimination, substantially surpassing all critic models.

\textbf{Generalization Across STEM Task Domains.}
While previous results show that our detector performs well on pure mathematical tasks, many real-world questions require reasoning in other STEM domains, such as science and finance. We therefore collect prompts from GPQA, OlympiadBench-Phy, and FinQA to construct a more comprehensive set of evaluation samples. Since our detector has already demonstrated superior performance over the two relatively weak critic models: R1-Qwen3-8B and QwQ-32B, we only report the performance of GPT-4.1 and o4-mini as a reference. As shown in Figure~\ref{fig:addition_bench_comparison}, our detector consistently achieves substantial improvements in hallucination detection over the backbone model across all benchmarks. Similar to the mathematical tasks, our detector outperforms GPT-4.1 and demonstrates competitive performance against o4-mini. This underscores the robustness of our approach as our detector generalizes effectively to diverse STEM domains despite being trained only on pure mathematical samples.

\subsection{Ablation Studies on Training Settings}
\begin{wrapfigure}{r}{0.58\textwidth}
\centering
\captionof{table}{Performance comparison between uniform and adaptive ensemble strategies. We report the average $S_{\textrm{incor}}$ / $S_{\textrm{cor}}$ of ensembled labels from different combinations of critic models.}
\label{tab:ensemble_comparison}
\resizebox{0.56\textwidth}{!}{
\begin{tabular}{lccc}
\toprule
\textbf{Ensemble Strategy} & \textbf{Comb 1} & \textbf{Comb 2} & \textbf{Comb 3} \\
\midrule
Uniform  & 60.21 / 100.0 & 66.75 / 99.99 & 69.97 / 99.98 \\
Adaptive & 78.16 / 99.99 & 78.76 / 99.99 & 78.77 / 99.99 \\
\bottomrule
\end{tabular}
}
\end{wrapfigure}
To understand how different training settings influence the detection performance, we conduct ablation studies on three aspects of the training recipe: the ensemble strategy for aggregating hallucination annotations (Section~\ref{subsec:token_ensemble}), the training objective (Section~\ref{subsec:detector_training}), and the scalability of training data. Throughout these experiments, we use Qwen3-8B as the backbone and Math as the training dataset.

\textbf{Adaptive Ensemble Consistently Improves Detection Performance.}
We first compare the effectiveness of two ensemble strategies for generating training labels: a simple uniform ensemble that averages the annotations, and an adaptive ensemble that learns weights for each critic model. To obtain the ensemble weights, we sample a small held-out subset from the training samples to form the validation set $\mathcal{D}_{\mathrm{val}}$ and minimize the loss in Eq.~\ref{eq:ensemble_loss} to learn weights for individual models. We then use these weights to aggregate the annotations into a single token-level label sequence for each training example. 
\begin{wrapfigure}{r}{0.5\textwidth}
    \vspace{-0.25cm}
    \centering
    \includegraphics[width=0.5\textwidth]{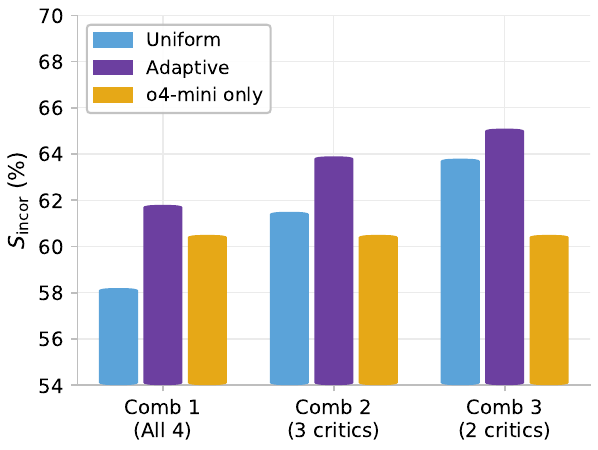}
    \caption{Detection performance under two ensemble strategies.}
    \vspace{-0.25cm}
    \label{fig:ensemble_strategy_abl}
\end{wrapfigure}
We start by evaluating the quality of the ensembled labels under three combinations. Comb 1 uses all four critic models, Comb 2 uses QwQ-32B, GPT-4.1, and o4-mini, and Comb 3 uses GPT-4.1 and o4-mini. We evaluate the ensembled token-level labels against the ground-truth labels produced by the labeler models. As shown in Table~\ref{tab:ensemble_comparison}, adaptive ensembling consistently improves label quality across all combinations, increasing the $S_{\textrm{incor}}$ score by 17.95, 12.01, and 8.8 points for Comb 1, 2, and 3, respectively, while maintaining $S_{\textrm{cor}}$. We then evaluate the impact of these ensemble strategies on actual model training. For comparison, we also train a detector using labels solely from the strongest critic model, o4-mini. In Figure~\ref{fig:ensemble_strategy_abl}, we report $S_{\textrm{incor}}$ while controlling all $S_{\textrm{cor}}$ at approximately 96 to ensure a fair comparison. The results indicate that detectors trained with adaptively ensembled data consistently perform better than the uniform counterpart. More importantly, we observe that detectors trained with data annotated by o4-mini alone underperform those using adaptive ensembles. This suggests that different models capture diverse hallucination patterns, and combining these annotations yields more robust training signals than relying on a single source.

\begin{wrapfigure}{r}{0.5\textwidth}
    \vspace{-0.2cm}
    \centering
    \includegraphics[width=0.5\textwidth]{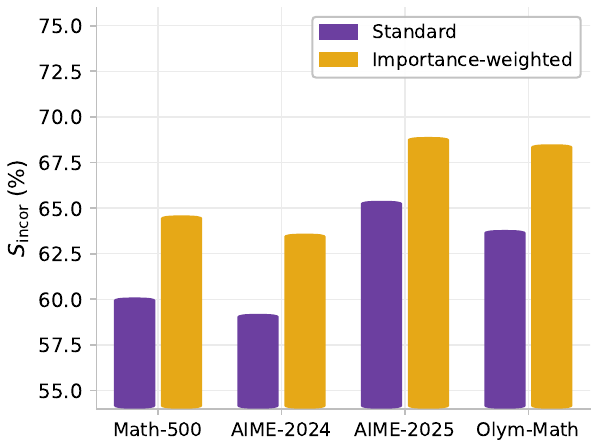}
    \caption{Detection performance under two training strategies.}
    \label{fig:training_loss_abl}
    \vspace{-0.2cm}
\end{wrapfigure}

\textbf{What if the Detector is Trained with Different Loss Objectives?}
We next explore how the training strategy affects detector performance. We consider two training schemes described in Section~\ref{subsec:detector_training}: standard training and importance-weighted training (Eq.~\ref{eq:weighted_loss}), which upweights tokens with high hallucination scores. We train detectors under both training schemes and evaluate them on four mathematical benchmarks. As shown in Figure~\ref{fig:training_loss_abl}, we report $S_{\textrm{incor}}$ ($S_{\textrm{cor}}$ is controlled around 96 to ensure a fair comparison) for two training strategies. Our results show that the importance-weighted strategy consistently improves detection performance across all benchmarks. These results suggest that our strategy is effective when hallucinated tokens are sparse, since upweighting hallucinated tokens helps mitigate the resulting class imbalance during training. We further study the impact of training data scaling on detection performance in Appendix~\ref{appendix:data_scaling}.

\section{Generalization of Hallucination Detection}
While our detector performs well on several mathematical and STEM tasks, real-world deployment still faces multiple challenges. In particular, the \emph{policy model} that produces the responses may differ from the one used to generate our training data, and user queries may come from non-mathematical \emph{task domains} where hallucination patterns may differ substantially. In the following sections, we evaluate our detector under these scenarios and explore simple strategies to enhance its generalization.

\subsection{Shifts in Policy Models and Task Domains}
\label{subsec:policy_task}
\textbf{Policy Models.} To evaluate the performance across different policy models, we generate evaluation samples with multiple policies on mathematical tasks. We consider two closed-source models, Gemini-2.0-Flash~\cite{google2024gemini_ai_update_december} and Claude-3.5-Haiku~\cite{anthropic_claude_sonnet_3_5}, and an open-source model, Qwen2.5-7B-Instruct. We report results on Math-500.

\begin{wrapfigure}{r}{0.55\textwidth}
    \centering
    \includegraphics[width=0.55\textwidth]{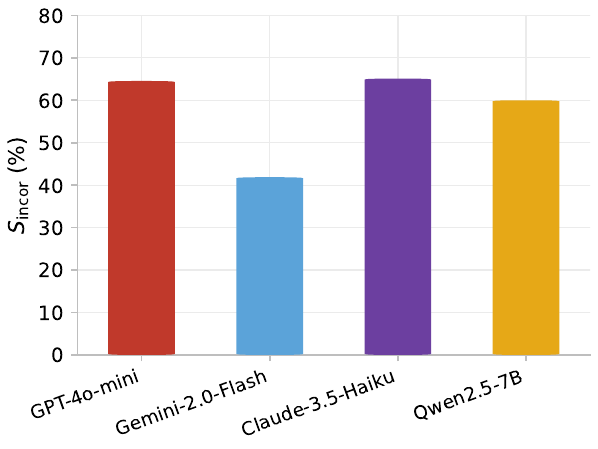}
    \caption{Detection performance across diverse policy models. The open-source policy model is Qwen2.5-7B-Instruct.}
    \label{fig:policy_model_abl}
\end{wrapfigure}

\textbf{Task Domains.} In addition to diverse policy models, real-world queries may also come from a wide range of task domains. Here, we consider a practical task, code generation. We evaluate on Code-Elo~\cite{quan2025codeelo} and LiveCodeBench-Lite~\cite{jain2024livecodebench} and generate evaluation samples using GPT-4o-mini and Gemini-2.0-Flash as policy models. To obtain high-quality hallucination annotations for coding tasks, we employ GPT-5 and Claude-4.5-Sonnet~\cite{anthropic_claude_sonnet_4_5} as our labeler models, and apply the uniform ensemble strategy for aggregating annotations.

\textbf{Preliminary Results on Generalization.} We begin with our ``baseline'' detector (baseline \textsc{TokenHD}-8B) trained only with mathematical data generated by GPT-4o-mini (discussed in Section~\ref{subsec:exp_settings}). For different policy models, Figure~\ref{fig:policy_model_abl} ($S_{\textrm{cor}}$ is controlled around 96 to ensure a fair comparison) shows that our detector demonstrates robust generalization across various policy models, including Claude-3.5-Haiku and Qwen2.5-7B-Instruct, even though our training data is solely sourced from GPT-4o-mini. Nevertheless, Gemini-2.0-Flash presents a more challenging distribution relative to our GPT-4o-mini training data, motivating the generalization strategies in Section~\ref{subsec:generalization}. For task-domain generalization, Table~\ref{tab:policy_model_comparison} shows that our baseline \textsc{TokenHD}-8B achieves 49.37 / 98.95 on Code-Elo and 41.30 / 98.70 on LiveCodeBench-Lite under the GPT-4o-mini policy. These results, obtained zero-shot with a math-only model on code tasks, highlight the need to expand training coverage to broader hallucination patterns. We investigate two practical strategies for this in the following section.

\subsection{Improving Detector's Generalization Capability}
\label{subsec:generalization}
To augment the training set, we collect additional samples that vary both the policy model and the task domain. For policy-model diversity, we use Gemini-2.0-Flash to generate additional mathematical training samples on the same Math dataset. For task-domain diversity, we collect code generation samples using the OpenCodeReasoning dataset~\cite{ahmad2025opencodereasoning} using both GPT-4o-mini and Gemini-2.0-Flash as policy models. In all cases, we sample two responses per prompt following our previous settings. To handle the newly added data more efficiently, we consider two strategies with different computational costs and compare their effectiveness. \textbf{Mix Training Data}: We aggregate training data from different sources and train our detector on this combined dataset. This strategy serves as a strong baseline, though it requires retraining the detector whenever new training data is updated, which incurs a high computational cost. \textbf{Model Merging:} We train separate specialized detectors for different domains and then merge their weights into a single detector. This enables modular updates and aggregates expertise from multiple specialized models without large-scale retraining. Specifically, we adopt several common methods from previous studies~\cite{wu2025unlocking}, including: average merging~\cite{wortsman2022model}, task vector~\cite{ilharco2022editing}, TIES-Merging~\cite{yadav2023ties} and DARE-Merging~\cite{yu2024language} (details can be accessed in Appendix~\ref{appendix:model_merging}). As shown in Table~\ref{tab:policy_model_comparison}, mix training on combined math and code data substantially improves code detection while maintaining math performance. Model merging can also improve generalization at relatively low computational cost. We also explore practical downstream applications of \textsc{TokenHD}, including best-of-N candidate selection and targeted self-correction, demonstrating further utility of token-level error scores beyond detection (see Appendix~\ref{appendix:applications}).

\begin{table*}[t]
\centering
\caption{Performance comparison between different strategies to improve the detector's generalization. We report $S_{\textrm{incor}}$ / $S_{\textrm{cor}}$.}
\label{tab:policy_model_comparison}
\resizebox{0.98\textwidth}{!}{
\begin{tabular}{l|ccc|ccc}
\toprule
\multirow{2}{*}{\textbf{Strategy}}
& \multicolumn{3}{c|}{\textbf{GPT-4o-mini}}
& \multicolumn{3}{c}{\textbf{Gemini-2.0-Flash}} \\
\cmidrule(lr){2-4} \cmidrule(lr){5-7}
& \textbf{Math-500} & \textbf{Code-Elo} & \textbf{LiveCode-Lite}
& \textbf{Math-500} & \textbf{Code-Elo} & \textbf{LiveCode-Lite} \\
\midrule
Baseline \textsc{TokenHD}-8B & 64.55 / 98.71 & 49.37 / 98.95 & 41.30 / 98.70 & 41.42 / 95.92 & 22.35 / 94.66 & 27.57 / 97.59 \\
\midrule
Mix Training           & 64.56 / 98.63 & 73.61 / 95.14 & 70.16 / 93.55 & 42.47 / 97.24 & 68.21 / 91.01 & 56.80 / 90.02 \\

Average Merging        & 48.75 / 93.53 & 68.00 / 91.29 & 63.53 / 87.74 & 33.42 / 61.09 & 67.23 / 91.14 & 51.88 / 87.80 \\
Task Vector            & 41.17 / 95.57 & 68.16 / 92.96 & 63.63 / 90.57 & 32.12 / 81.04 & 65.23 / 92.84 & 48.08 / 91.86 \\
TIES-Merging           & 50.47 / 91.47 & 67.84 / 90.81 & 63.35 / 87.24 & 33.61 / 56.34 & 67.25 / 89.76 & 52.14 / 85.80 \\
DARE-Merging           & 49.25 / 91.77 & 67.58 / 90.89 & 63.43 / 87.70 & 32.94 / 26.88 & 66.78 / 89.97 & 51.93 / 86.66 \\
\bottomrule
\end{tabular}
}
\end{table*}

\section{Related Work}

\textbf{Hallucination Detection and Mitigation.} Despite the exceptional capabilities of LLMs, they still suffer from hallucinations~\cite{huang2025survey, kalai2025language, he2025mmboundary}, where the generated text appears fluent but contains errors within the reasoning process. Many prior studies primarily investigate factual hallucinations, where LLM-generated contexts conflict with real-world knowledge or include unverifiable claims. To mitigate these issues, various strategies have been proposed, such as leveraging specialized decoding strategies~\cite{wang2024mitigating} or RAG systems~\cite{gao2023retrieval, lewis2020retrieval} to prevent hallucination during inference, fine-tuning policy models to mitigate hallucinations~\cite{zhang2024r, zhang2025law, he2026empowering}, and post-hoc strategies to detect hallucination after generation~\cite{manakul2023selfcheckgpt, chen2023hallucination, miao2023selfcheck, sriramanan2024llm, obeso2025real}. In contrast, we study hallucinations in reasoning-intensive tasks, where models generate contents that appear coherent but contain logical flaws, leading to incorrect final answers. Moreover, existing detection studies often rely on synthesized hallucinations, resulting in limited hallucination patterns, and may not reflect models’ behaviors in practice. We avoid manually crafted hallucinations and instead build an automated data engine that samples reasoning traces from policy models and annotates token-level hallucinations using our scalable annotation framework, capturing more authentic hallucination behaviors beyond a fixed set of curated patterns.

\textbf{Reward Models for LLMs.}
Reward models (RMs) are critical in defining preference criteria and shaping the generation quality of LLMs. Conventional RMs typically assign a numerical score to an entire sequence to rank candidate responses~\cite{zhu2024starling,liu2024skywork,zhong2025comprehensive,liu2024rm, liu2025pairjudge}, yet they provide sparse feedback and struggle to identify where errors occur within the reasoning. To address this, the research community proposed Process Reward Models (PRMs) to supervise intermediate reasoning steps~\cite{lightman2023let, wang2024math, song2025prmbench, zhang2025lessons, zheng2025processbench}. However, existing PRMs depend on heuristic step boundaries that are hard to define for free-form text or require intensive MCTS~\cite{browne2012survey} sampling, incurring high computational overhead. In contrast, \textsc{TokenHD} provides dense reward supervision on free-form content without requiring step segmentation or costly MCTS sampling.

\section{Conclusion and Limitations}
In this paper, we present \textsc{TokenHD}, a holistic framework for developing fine-grained hallucination detectors in reasoning-intensive tasks. \textsc{TokenHD} establishes a complete pipeline, integrating a scalable data engine for synthesizing high-quality annotations with an importance-weighted training strategy to mitigate label sparsity. Unlike conventional PRMs, our detector operates directly on free-form text at the token level, enabling precise hallucination localization without requiring predefined step separation. To validate the effectiveness of \textsc{TokenHD}, we establish a rigorous evaluation framework and conduct extensive experiments on mathematical and STEM benchmarks. Experimental results demonstrate that our lightweight detector can surpass much larger reasoning models while maintaining a significantly lower inference overhead. In sum, \textsc{TokenHD} offers a highly scalable and cost-efficient solution for fine-grained hallucination detection, providing a practical tool for enhancing the reliability of LLMs in complex reasoning.

\bibliographystyle{unsrt}
\bibliography{example_paper}

\newpage
\appendix
\onecolumn

\section{Broader Impacts}
This paper introduces \textsc{TokenHD}, a framework designed for token-level hallucination detection in LLM-generated responses. Our trained detector is highly efficient and effective in pinpointing subtle hallucinations within reasoning-intensive tasks. Since our detector is lightweight and inference-efficient, it can be seamlessly integrated into most existing LLM systems to detect the potential hallucinations of generated text without significant computational overhead. In sum, our detector contributes to enhancing the truthfulness of LLM-generated content, which is a crucial step toward the reliable deployment of LLMs in real-world applications.

\section{A Closer Look at How We Obtain the Token-level Hallucination Annotations}
\label{appendix:data_annotation}
\subsection{Prompts for Data Annotation}
This section describes the prompt designs used in our data annotation process. To identify hallucinated text within generated responses, we prompt LLMs to achieve this. Specifically, we leverage advanced and highly capable models as labeler models to obtain ground-truth annotations for evaluation. We use relatively cheaper or open-source models as critic models to generate large-scale training data due to cost considerations. We apply the same annotation prompts for both labeler and critic models. Our experiments primarily involve three categories of data sources, including the mathematical tasks: Math training data, AceReason-Math, Big-Math, Math-500, AIME-2024, AIME-2025 and Olym-Math, STEM tasks: Olym-Phy, GPQA and FinQA and code generation tasks: OpenCodeReasoning, Code-ELO and LiveCodeBench-Lite. For both mathematical and STEM tasks, we use the prompt shown in Figure~\ref{fig:math_prompt}. For code generation tasks, we use the prompt in Figure~\ref{fig:code_prompt}.

\subsection{Details of Text Restoration}
After obtaining the textual annotations, we map these text fragments back to the token space. This process requires each identified fragment to match its corresponding part in the original response exactly. However, we found that the raw text from the LLM often differs slightly from the original text, typically in formula formatting or paragraph breaks. To address this, we prompt o4-mini to restore the raw text (the prompt is shown in Figure~\ref{fig:restore_prompt}) so that each fragment can be correctly aligned with the original response. Since a single restoration is often insufficient, we developed an iterative restoration strategy, detailed in Algorithm~\ref{alg:text_restoration}. This method effectively restores fragments to their original text with high efficiency. In our experiments, we set the max iteration rounds to 3. Table~\ref{tab:restoration_rate} shows the direct-match rate (fraction of raw critic spans that already match the original text verbatim) and the post-restoration rate (fraction successfully aligned after iterative restoration). On average across all four critics, the restoration process recovers 98.10\% of spans, a large improvement over the 64.25\% direct-match baseline.

\begin{table}[h]
\centering
\caption{Text restoration rates across four critic models. Our algorithm largely improves the match rate of text spans after restoration.}
\label{tab:restoration_rate}
\setlength{\tabcolsep}{6pt}
\begin{tabular}{lcc}
\toprule
\textbf{Critic} & \textbf{Direct Match (\%)} & \textbf{Post-Restoration (\%)} \\
\midrule
DeepSeek-R1-Qwen3-8B & 26.57 & 96.43 \\
QwQ-32B              & 44.18 & 97.46 \\
GPT-4.1              & 95.66 & 99.45 \\
o4-mini              & 83.19 & 98.74 \\
\midrule
\textbf{Average}     & \textbf{64.25} & \textbf{98.10} \\
\bottomrule
\end{tabular}
\end{table}

\subsection{How to Filter High-quality Data}
Before training our detector, we apply a filtering strategy to improve the quality of the dataset. First, we remove samples with incomplete or corrupted generation traces. For the remaining data, we filter out samples that have incorrect final results but contain no identified hallucinations. We also remove samples with low annotation consistency. Specifically, a sample is removed from our dataset if the maximum aggregated hallucination score is below 0.5.

\section{Details of Training Hallucination Detector}
\label{appendix:training_settings}
For generating mathematical training samples, we leverage the whole Math training dataset and subsets of AceReason-Math and Big-Math, obtaining around 49,000 valid samples after data filtering. To balance the dataset, we supplement it with a small set of non-hallucinated samples (those with correct answers and no identified errors). We train for one epoch using the Adam optimizer~\cite{kingma2014adam} with a cosine decay schedule and linear warmup, setting the peak learning rate to $1 \times 10^{-5}$.

\section{Ground-Truth Label Quality Verification}
\label{appendix:localization}

\subsection{Annotation Rubric}
\label{appendix:rubric}

Table~\ref{tab:rubric} lists the rubric provided to human annotators for identifying and marking hallucinated spans in model-generated solutions. The criteria are applied in order; later criteria address ambiguous edge cases that arise in complex multi-step reasoning.

\begin{table}[h]
\centering
\caption{Rubric for hallucination span annotation. Human annotators apply these criteria in order to identify and mark erroneous spans in model-generated solutions.}
\label{tab:rubric}
\resizebox{\textwidth}{!}{
\begin{tabular}{cp{3.0cm}p{9.8cm}}
\toprule
\textbf{\#} & \textbf{Criterion} & \textbf{Guideline} \\
\midrule
R1 & Error definition &
Mark a span as hallucinated if it contains: (a) an incorrect mathematical operation or numerical result; (b) a logically invalid inference (e.g., incorrectly applying a theorem, reversing an implication, or making an unsupported leap); (c) a factual claim not supported by the problem statement or standard mathematical knowledge; or (d) an incorrect intermediate conclusion that directly propagates error to subsequent reasoning steps or the final answer. \\
\addlinespace
R2 & Span granularity &
Mark the \emph{minimal} contiguous span that contains the error; do not extend the marked span to include surrounding correct text. If the error involves a single symbol, value, or expression, mark the smallest clause or equation that makes the error meaningful and interpretable in context. \\
\addlinespace
R3 & Multiple errors &
When multiple distinct errors exist, mark each independently using separate, non-overlapping spans. If error A causes a downstream error B, mark both independently. Do not merge errors separated by correct intervening text, even if they are logically related. \\
\addlinespace
R4 & Correct solutions &
If the solution is entirely correct and reaches the correct final answer, output ``No errors!''. Do not mark minor stylistic differences, equivalent reformulations of correct expressions, or unnecessarily verbose but correct steps. \\
\addlinespace
R5 & Ambiguous steps &
Do not mark a step that is redundant but mathematically valid. When a step follows from an earlier erroneous assumption, prioritize marking the original erroneous assumption as the primary error span; mark the downstream step only if it introduces an \emph{additional} independent error beyond what was already caused by the upstream mistake. \\
\bottomrule
\end{tabular}
}
\end{table}

\subsection{Annotation Quality Assessment}
\label{appendix:annotation_quality}

To verify that the rubric produces reliable annotations in practice, we recruited human annotators to rate a held-out set of annotated samples on a 1--5 scale along two axes: \textbf{Accuracy} (whether the marked spans actually contain errors) and \textbf{Completeness} (whether the major errors in the solution are covered). Annotators followed the rubric above and were assisted by an advanced LLM (Gemini-3.1-Pro) to help verify their judgments and improve rating consistency; final scores reflect human decisions.

\begin{figure}[h]
\centering
\includegraphics[width=0.46\textwidth]{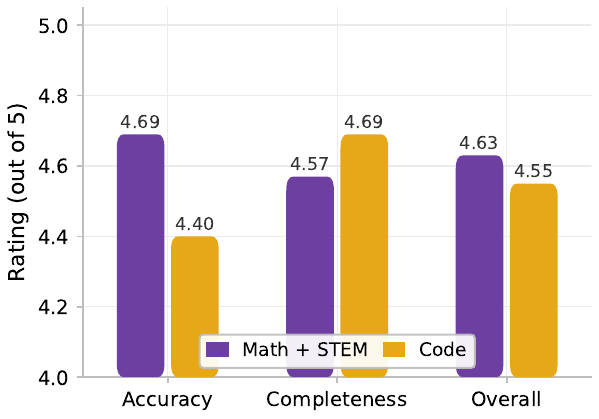}
\caption{Human annotation quality assessment (1--5 scale). Annotators rated GT annotations on accuracy and completeness across two task domains, assisted by an advanced LLM for verification.}
\label{fig:annotation_quality}
\end{figure}

As shown in Figure~\ref{fig:annotation_quality}, overall scores are high across all task domains (math + STEM: 4.63/5, code: 4.55/5). Accuracy is particularly strong for math and STEM tasks (4.69), indicating that annotated spans are rarely false positives. Completeness (4.57) is modestly lower, consistent with the difficulty of exhaustively identifying all errors in complex multi-step solutions. For code tasks, the pattern inverts (accuracy 4.40, completeness 4.69), which we attribute to token-boundary ambiguity in code blocks rather than incorrect error identification. Taken together, these scores confirm that our rubric-guided annotation pipeline produces high-quality token-level supervision.

\section{Extended Discussions of our \textsc{TokenHD} Framework}
\label{appendix:extended_discussion}
\subsection{Relationship between our Detector and Reward Models}
Reward models are typically designed to provide training signals that reflect specific preferences, allowing subsequent training procedures such as reinforcement learning to select better responses and improve the policy model's performance. While our work shares a similar form with reward models, we differ in our objectives. Our detector is specifically designed to identify hallucinations at the token level, serving as an indicator of where hallucinations occur. Despite this difference in usage, our paradigm could potentially provide dense, token-level rewards, offering more fine-grained supervision for training and ranking responses than existing response-level or step-level signals.

\subsection{Ablation of Training Settings}
Our primary experiments utilize the Qwen3 series as backbone models, with a focus on small-scale architectures. We make these choices for two key considerations. First, we observe that training with a larger backbone, such as Qwen3-14B, yields an improvement of approximately 2 points in $S_{\textrm{incor}}$ over the 8B version on the Math-500 dataset. As detailed in Section~\ref{subsec:exp_settings} and Table~\ref{tab:math_benchmark}, performance scales consistently with model size across this range. Even the smallest 0.6B variant achieves 63.64 on Math-500, outperforming QwQ-32B (55.05) by a substantial margin. We also use other model families as backbones, such as Qwen2.5 and Llama3, and find that the Qwen3 series consistently performs better. Second, we prioritize efficiency to ensure the detector is lightweight for practical deployment. Since our detector is primarily designed to serve as a plug-in hallucination detection module following much larger LLM systems, using a small detector is critical to maintain high inference efficiency and minimize additional computational overhead.

\subsection{Impact of Training Data Scaling}
\label{appendix:data_scaling}
\begin{figure}[h]
    \centering
    \includegraphics[width=0.5\textwidth]{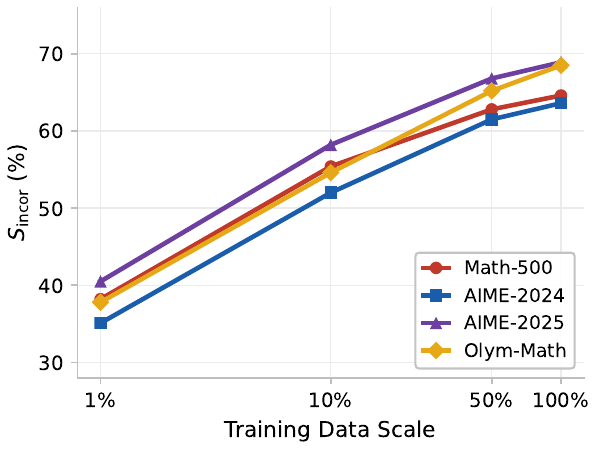}
    \caption{Detection performance across different training data scales on mathematical tasks.}
    \label{fig:scaling_curve}
\end{figure}

To assess the impact of data scale, we train our detectors on subsets of the training set sampled at 1\%, 10\%, 50\%, and 100\% and evaluate them on the four mathematical benchmarks. As shown in Figure~\ref{fig:scaling_curve}, increasing the data size consistently improves $S_{\textrm{incor}}$ until reaching around 50\% of the training dataset across all benchmarks. After this point, we observe that the improvement in detection performance slows down significantly, indicating that the gains from adding more training data start to saturate.

\subsection{Advantages Compared to Larger Reasoning Models}
In our evaluation, we observe that our detector significantly outperforms its original backbone models in hallucination detection, even though those backbones provide much more reasoning content. We also find that our detector can outperform larger, more advanced critic models, despite being trained on data labeled by them. While we acknowledge that o4-mini still leads on Math-500, our approach offers two distinct advantages. First, our model is much more lightweight than o4-mini, leading to faster inference speeds and significantly lower API costs. Second, our detector does not rely on text restoration, which can directly identify hallucinated text fragments within the original response. In contrast, o4-mini requires an extra text restoration step to extract exact fragments, which adds computational cost and complexity to the pipeline. With only a modest gap in detection performance, our model provides a much simpler and more efficient solution for hallucination detection in practice.

\subsection{Different Roles between Critic and Labeler Models}
Our methodology relies on two sets of models for hallucination annotation: critic models for generating training data and a labeler model for producing ground-truth labels for evaluation. Although these models follow a similar annotation workflow and use identical prompts, they serve distinct purposes. First, there is a trade-off regarding annotation costs. Advanced models like GPT-5 are extremely expensive, making them less practical for generating large-scale training data. Therefore, we utilize a set of more cost-effective critic models for training while reserving the most powerful labeler for annotating evaluation samples. Second, employing distinct model sets for annotating training and evaluation data prevents direct model distillation. This ensures a more rigorous and fair assessment. Notably, although our detector is trained on data annotated by the critic models, it outperforms most of these critic models themselves. This demonstrates that our approach goes beyond simple imitation. Instead, it provides an efficient and robust training framework that effectively identifies hallucinations at the token level.

\section{Comparison with Process Reward Models}
\label{appendix:prm_comparison}

Process Reward Models (PRMs) assign correctness scores to each reasoning step, making them the closest existing paradigm to our token-level hallucination detection task. We compare \textsc{TokenHD} against two strong open-source PRM baselines, Qwen2.5-Math-PRM-7B and Qwen2.5-Math-PRM-72B~\cite{prmqwen25math}, designed for mathematical reasoning. To enable a fair token-level comparison, we adapt each PRM as follows: we first obtain the step-level correctness score for each reasoning step produced by the PRM, and then uniformly distribute this score to every token within that step, yielding a token-level error probability sequence. This adaptation is the most direct way to convert step-level signals to token-level predictions; more adaptation would require additional modification beyond the PRM's original design. These sequences are evaluated against the same ground-truth labels used throughout the paper (annotated by o3 and GPT-5 under the same evaluation protocol).

As shown in Table~\ref{tab:prm_comparison}, \textsc{TokenHD}-8B outperforms Qwen2.5-Math-PRM-72B by 30 to 57 points across all five benchmarks. We attribute this gap to two key factors. First, PRMs are trained to rank steps as correct or incorrect at a coarse granularity: even when a step contains a critical hallucination, the PRM may assign it a moderate score because the surrounding steps appear plausible. The resulting token-level scores are therefore a blunt proxy for fine-grained error localization. Second, PRMs inherently depend on explicit step segmentation, which constrains their resolution to predefined structural boundaries. Our detector, by contrast, is trained end-to-end to predict per-token hallucination scores directly on free-form text, enabling it to identify errors within individual steps without any structural assumptions. These results underscore the importance of optimizing specifically for token-level detection rather than adapting step-level correctness signals from a coarser annotation source.

\begin{table}[h]
\centering
\caption{Token-level $S_{\textrm{incor}}$ comparison between PRMs and \textsc{TokenHD}-8B. For a fair comparison, PRM step scores are uniformly distributed to each token within the step. GT labels: o3 + GPT-5. Policy: GPT-4o-mini.}
\label{tab:prm_comparison}
\resizebox{0.85\textwidth}{!}{
\begin{tabular}{lccccc}
\toprule
\textbf{Model} & \textbf{Math-500} & \textbf{Olym-Math} & \textbf{GPQA} & \textbf{Olym-Phy} & \textbf{FinQA} \\
\midrule
Qwen2.5-Math-PRM-7B  & 26.25 & 18.41 & 5.17 & 8.29 & 8.35 \\
Qwen2.5-Math-PRM-72B & 34.58 & 21.62 & 8.14 & 13.77 & 13.69 \\
\textsc{TokenHD}-8B  & \textbf{64.55} & \textbf{68.45} & \textbf{65.49} & \textbf{65.85} & \textbf{64.35} \\
\bottomrule
\end{tabular}
}
\end{table}

\section{AUROC and AUPRC}
\label{appendix:auroc}

Beyond the primary $S_{\textrm{incor}}$ and $S_{\textrm{cor}}$ metrics, we report AUROC and AUPRC to provide a threshold-independent view of detection performance. Both metrics are computed at the token level on incorrect samples only. For each benchmark, we pool all tokens from all incorrect samples and form a set of (predicted score, ground-truth label) pairs, where the predicted score is the continuous output $\widehat{s}_i \in [0,1]$ from the detector and the ground-truth label is the binarized annotation $\mathbb{I}[s_i > \beta_I]$. AUROC is then computed as the area under the ROC curve over this pooled set, measuring the probability that a randomly chosen hallucinated token receives a higher predicted score than a randomly chosen non-hallucinated token. AUPRC is the area under the precision-recall curve over the same pooled set, and is more informative in our setting because hallucinated tokens are sparse within a response: the precision-recall curve directly captures the trade-off between localization precision and recall at every operating point, and is more sensitive to the detection of rare positive tokens than AUROC.

For critic models, the predicted score for each token is its binary annotation: a token inside a flagged error span receives a score of 1, and all other tokens receive a score of 0. This binary scoring is used directly to compute AUROC and AUPRC for the critics. The asymmetry in score type (critics output binary values while \textsc{TokenHD} outputs continuous scores) reflects a genuine capability difference rather than an unfair evaluation design. Table~\ref{tab:auroc_comparison} reports results for \textsc{TokenHD}-8B and four critic models on four mathematical benchmarks. \textsc{TokenHD}-8B achieves the best AUROC on all four benchmarks, ranging from 0.8739 to 0.9004, surpassing the strongest critic per benchmark by 0.04--0.09.

This advantage stems from a fundamental difference in the nature of the two approaches' outputs. Critic models produce hard binary annotations: a token is either inside a flagged span or not, leaving no room for confidence or gradation. When a critic misses an error entirely, the predicted score for every token in that sample collapses to zero. \textsc{TokenHD}, by contrast, outputs continuous scores for every token, assigning high values to likely hallucinated positions and near-zero values elsewhere. This graded output gives a richer signal for ranking tokens, which is directly reflected in the higher AUROC and AUPRC values.

\begin{table}[h]
\centering
\caption{AUROC and AUPRC at the token level on four mathematical benchmarks.}
\label{tab:auroc_comparison}
\resizebox{\textwidth}{!}{
\begin{tabular}{llcccc}
\toprule
\textbf{Metric} & \textbf{Model} & \textbf{Math-500} & \textbf{Olym-Math} & \textbf{AIME-2024} & \textbf{AIME-2025} \\
\midrule
\multirow{5}{*}{AUROC}
 & \textsc{TokenHD}-8B & \textbf{0.8877} & \textbf{0.9004} & \textbf{0.8739} & \textbf{0.8947} \\
 & o4-mini             &          0.8489 &          0.8288 &          0.7880 &          0.8226 \\
 & GPT-4.1             &          0.7042 &          0.7591 &          0.7206 &          0.7543 \\
 & QwQ-32B             &          0.7358 &          0.7283 &          0.6904 &          0.7082 \\
 & R1-Qwen3-8B         &          0.6850 &          0.6703 &          0.6484 &          0.6435 \\
\midrule
\multirow{5}{*}{AUPRC}
 & \textsc{TokenHD}-8B & \textbf{0.7537} & \textbf{0.7760} & \textbf{0.7359} & \textbf{0.8014} \\
 & o4-mini             &          0.6910 &          0.6423 &          0.5908 &          0.6088 \\
 & GPT-4.1             &          0.4271 &          0.4878 &          0.4426 &          0.4987 \\
 & QwQ-32B             &          0.5545 &          0.5292 &          0.4919 &          0.5212 \\
 & R1-Qwen3-8B         &          0.5031 &          0.4696 &          0.4562 &          0.4663 \\
\bottomrule
\end{tabular}
}
\end{table}

\section{Details of Model Merging}
\label{appendix:model_merging}

Let $\theta_{\mathrm{math}}$ and $\theta_{\mathrm{code}}$ denote the parameters of the math-only and code-only detectors, and $\theta_{\mathrm{base}}$ the shared Qwen3-8B backbone. We define the task vectors as $\tau_{\mathrm{math}}=\theta_{\mathrm{math}}-\theta_{\mathrm{base}}$ and $\tau_{\mathrm{code}}=\theta_{\mathrm{code}}-\theta_{\mathrm{base}}$. We also denote the classification head (which maps the final hidden state to token-level hallucination probabilities) as $\theta_{\mathrm{head}}$, which is the most calibration-sensitive component of the model.

\begin{itemize}[leftmargin=1.5em,itemsep=1pt,topsep=2pt]
\item \textbf{Average Merging}: $\theta_{\mathrm{merged}} = \tfrac{1}{2}(\theta_{\mathrm{math}}+\theta_{\mathrm{code}})$.
\item \textbf{Task Vector}: $\theta_{\mathrm{merged}} = \theta_{\mathrm{base}} + \alpha(\tau_{\mathrm{math}}+\tau_{\mathrm{code}})$, $\alpha=1.0$.
\item \textbf{TIES-Merging}: For each parameter dimension, we prune entries of $\tau_{\mathrm{math}}$ and $\tau_{\mathrm{code}}$ with absolute value below the top-20\% threshold (i.e., retaining the 20\% largest-magnitude entries), resolve sign conflicts between the two pruned vectors by majority vote, and add the merged result back to $\theta_{\mathrm{base}}$, $\alpha=1.0$.
\item \textbf{DARE-Merging}: Randomly drop 80\% of entries in $\tau_{\mathrm{code}}$, rescale survivors by $\frac{1}{1-0.8}$ to preserve expectation, and merge the result with $\theta_{\mathrm{math}}$ using task arithmetic: $\theta_{\mathrm{merged}} = \theta_{\mathrm{math}} + \alpha\,\tilde{\tau}_{\mathrm{code}}$, $\alpha=1.0$.
\end{itemize}

\begin{algorithm}[tb]
    \caption{Iterative Text Restoration}
    \label{alg:text_restoration}
    \begin{algorithmic}[1]

        \makeatletter
        \renewcommand{\COMMENT}[1]{\hfill$\triangleright$ #1}
        \makeatother

        \STATE {\bfseries Input:} Hallucinated text fragments $( r_m )_{m=1}^{M}=[r_1,\dots,r_M]$; Restoration Model $\mathcal{R}(\cdot)$; Max iteration rounds $N$; Original text response $\mathbf{y}$.
        \STATE {\bfseries Output:} List of successfully restored fragments $\mathbf{r}_{\text{res}}$.

        \STATE \textbf{Initialization:} $\mathbf{r}_{\text{res}} \leftarrow []$, $\mathbf{r}_{\text{unres}}^{0} \leftarrow (r_m)_{m=1}^{M}$.
        \FOR{$n = 1$ {\bfseries to} $N$}
            \STATE \textbf{Step 1: Restore text candidates}
            \STATE $\widetilde{\mathbf{r}}^{n} \leftarrow \mathcal{R}(\mathbf{r}_{\text{unres}}^{n-1}, \mathbf{y})$

            \STATE \textbf{Step 2: Verify the success of restoration}
            \STATE $\mathbf{r}_{\text{unres}}^{n} \leftarrow []$

            \FOR{each fragment $\widetilde{r} \in \widetilde{\mathbf{r}}^{n}$}
                \IF{$\widetilde{r}$ is contained within $\mathbf{y}$}
                    \STATE Append $\widetilde{r}$ to $\mathbf{r}_{\text{res}}$
                \ELSE
                    \STATE Append $\widetilde{r}$ to $\mathbf{r}_{\text{unres}}^{n}$
                \ENDIF
            \ENDFOR

            \STATE \textbf{Step 3: Early stopping when all text fragments are restored}
            \IF{$\mathbf{r}_{\text{unres}}^{n}$ is empty}
                \STATE \textbf{break}
            \ENDIF
        \ENDFOR

        \STATE {\bfseries return} $\mathbf{r}_{\text{res}}$
    \end{algorithmic}

\end{algorithm}

\begin{figure}[ht]

\begin{AIBox}{Prompt for Identifying Hallucinations in Math and STEM Tasks}

\parbox[t]{\textwidth}{

\scriptsize \begin{alltt}
 You are given a question and its corresponding solution. Your task is to analyze
   whether the solution completely and correctly solves the given problem.


**Your Tasks**: 

- Carefully examine the solution step by step

- Identify if there are any errors in the reasoning, logic, facts, calculations, or
  final result

- If errors are found, extract the exact erroneous parts from the solution
  and list them using error tags


**Instructions**:

- **You MUST identify all distinct errors in the solution. Do not stop
  after finding the first error.**

- Analyze the solution thoroughly for factual errors, mathematical errors, logical
  flaws, incorrect reasoning or wrong final answers.

- **If the solution contains any errors, you MUST extract and tag every
  erroneous text segment you find.**

- **IMPORTANT**: Use the original text from the solution without any
  modifications - maintain the exact same format, wording, punctuation,
  and mathematical notation as in the original.

- Keep the extracted error text **complete** and **relevant** to the
  specific error. Do **NOT** extract only mathematical symbols or formulas.

- Tag each error using `<error n></error n>` format, starting from
  index 1 and numbering sequentially.

- If the solution is completely correct, simply state "No errors!".


**Input Format**:

Problem: [PROBLEM\_STATEMENT]

Solution: [SOLUTION\_TEXT]


**Output Format**:

If errors are found:


<error 1>[EXACT\_TEXT\_FROM\_SOLUTION\_1]</error 1>

<error 2>[EXACT\_TEXT\_FROM\_SOLUTION\_2]</error 2>

<error 3>[EXACT\_TEXT\_FROM\_SOLUTION\_3]</error 3>

...


If no errors are found, simply output:

No errors!


Please analyze the following:


Problem: \{problem\}


Solution: \{solution\}
\end{alltt}}

\end{AIBox}
\caption{Prompts used for identifying hallucinations in mathematical and STEM tasks.}
\label{fig:math_prompt}
\end{figure}

\begin{figure}[ht]

\begin{AIBox}{Prompt for Identifying Hallucinations in Code Generation Tasks}

\parbox[t]{\textwidth}{

\scriptsize \begin{alltt}
 You are given a coding problem and its corresponding solution. Your task
   is to analyze whether the solution completely and correctly solves the
   given problem.


**Your Task**:

- Carefully examine the solution step by step, including the code
  implementation and any accompanying explanation.

- Identify if there are any errors in the syntax, logic, algorithm,
  implementation, or final result.

- If errors are found, extract the exact erroneous parts from the solution
  and list them using error tags.


**Instructions**:

- **You MUST identify all distinct errors in the solution. Do not stop
  after finding the first error.**

- Treat as an error any of the following: syntax or runtime errors, logical flaws or
  incorrect algorithms, wrong or incomplete handling of edge cases (e.g., empty
  inputs, boundary values defined by the problem), violations of the problem’s
  constraints (e.g., required input/output format, time/space limits), obviously
  incorrect or incomplete implementation of the required logic, misuse or
  misunderstanding of encodings, data types, or return values.

- Verify that all built-in functions and libraries are used correctly and
  that their behavior matches the problem requirements.

- Analyze both the code and any explanation text; contradictions with the
  problem or with the code are also errors.

- **If the solution contains any errors, you MUST extract and tag every
  erroneous text segment you find.**

- **Use only original text from the solution**: do not modify wording,
  formatting, punctuation, or indentation; do not invent any text.

- Keep each extracted error segment complete and relevant to that specific
  error. Do not extract only single symbols or isolated keywords unless
  they alone form the entire error.

- If one logical error spans multiple lines, you may include the entire
  block in a single error tag.

- Tag each error using `<error n></error n>` format, starting from 1
  and numbering sequentially.

- If the solution is completely correct, output exactly: "No errors!".


**Input Format**:

Problem: [PROBLEM\_STATEMENT]

Solution: [SOLUTION\_TEXT]


**Output Format**:

If errors are found:


<error 1>[EXACT\_TEXT\_FROM\_SOLUTION\_1]</error 1>

<error 2>[EXACT\_TEXT\_FROM\_SOLUTION\_2]</error 2>

<error 3>[EXACT\_TEXT\_FROM\_SOLUTION\_3]</error 3>

...


If no errors are found, simply output:

No errors!


Please analyze the following:


Problem: \{problem\}


Solution: \{solution\}
\end{alltt}}

\end{AIBox}
\caption{Prompts used for identifying hallucinations in code generation tasks.}
\label{fig:code_prompt}
\end{figure}

\begin{figure}[ht]

\begin{AIBox}{Prompt for Text Restoration}

\parbox[t]{\textwidth}{

\scriptsize \begin{alltt}
 You are given one original text and multiple extracted text segments that were
   supposedly extracted from the original text, but may contain slight variations,
   errors, or modifications.
 
1. **Original Text**: A long document or passage

2. **Extracted Text**: A segment supposedly extracted from the original
  text, but may contain slight variations, errors, or modifications


**Your Task**:

For each extracted text segment, locate the corresponding section in the
  original text that best matches it, and output the exact original text
  segment with appropriate tags.

**Instructions**:

- The extracted text may have minor differences from the original, such
  as: spelling variations or typos, punctuation differences, minor word
  substitutions, and formatting changes.

- Find the most similar section in the original text for each extracted
  segment. If multiple potential matches exist, choose the one with the
  highest similarity.

- Only return the portion of the original text that corresponds to the
  extracted segment. Do NOT add extra content before or after the matching
  section, and do NOT remove any content that should be included.

- **IMPORTANT**: Return the **EXACT** text from the original document
  that corresponds to each extracted segment. Pay attention to differences,
  such as *whitespace*, *punctuation*, *brackets*, and *line breaks*, etc.

- If no reasonable match is found, output "NO\_MATCH\_FOUND".


**Input Format**:

Original Text: [ORIGINAL\_TEXT]

<extract1>[EXTRACTED\_TEXT\_1]</extract1>

<extract2>[EXTRACTED\_TEXT\_2]</extract2>

<extract3>[EXTRACTED\_TEXT\_3]</extract3>

...


**Output Format**:

<result1>[EXACT\_SEGMENT\_FROM\_ORIGINAL\_1]</result1>

<result2>[EXACT\_SEGMENT\_FROM\_ORIGINAL\_2]</result2>

<result3>[EXACT\_SEGMENT\_FROM\_ORIGINAL\_3]</result3>

...


Please process the following input:

Original Text: \{original\_text\}

Extracted Text: \{extracted\_text\}

\end{alltt}}

\end{AIBox}
\caption{Prompts used for restoring the identified hallucinated text to match the original response.}
\label{fig:restore_prompt}
\end{figure}

\section{Applications: Best-of-N Selection and Self-Correction}
\label{appendix:applications}

We demonstrate two practical downstream applications of \textsc{TokenHD}: using token-level error scores as a scoring function for best-of-N candidate selection, and providing fine-grained error hints for self-correction. Both experiments are conducted on Math-500 with GPT-4o-mini as the policy model, focusing on samples where the initial response is incorrect.

\subsection{Best-of-N Selection}
\label{appendix:bon}

A natural application of a hallucination detector is to score multiple candidate solutions and select the most reliable one. For each incorrect sample, we generate ten candidate solutions and use \textsc{TokenHD}-8B to assign a per-token error probability to each candidate, aggregate these probabilities into a single candidate-level score, and select the candidate with the lowest aggregated score (i.e., the candidate with the least predicted hallucination). We compare three aggregation strategies: (1) \textbf{Full-Response Mean}: average token error probability across the entire response; (2) \textbf{Full-Response Min}: minimum token error probability across the response; (3) \textbf{Worst-10\% Mean}: average error probability of the 10\% of tokens with the highest scores, capturing the worst predicted region. We measure selection accuracy relative to the oracle ceiling of 56.4\%, the fraction of samples where at least one of the ten candidates is correct.

As shown in Table~\ref{tab:bon}, using the Full-Response Mean improves accuracy from the majority-vote baseline of 25.8\% to 31.8\%. By restricting aggregation to the minimum token score, Full-Response Min achieves the best result at 35.2\%, reaching 62.4\% of the oracle ceiling. This finding indicates that a response's most reliable token (the one with the lowest predicted error probability) provides a particularly discriminative signal for ranking candidates. Worst-10\% Mean (34.3\%) also substantially outperforms full-sequence averaging, further corroborating this observation. These results demonstrate that \textsc{TokenHD} can serve as an effective scoring function for best-of-N selection without any additional training or reward modeling.

\begin{table}[h]
\centering
\caption{Best-of-N selection results on incorrect Math-500 samples using \textsc{TokenHD}-8B as the scorer. Oracle ceiling = 56.4\% (fraction where at least one of ten candidates is correct). Majority vote is the baseline.}
\label{tab:bon}
\begin{tabular}{lcc}
\toprule
\textbf{Strategy} & \textbf{Accuracy} & \textbf{\% of Oracle} \\
\midrule
Majority vote (baseline) & 25.8\% & 45.7\% \\
Full-Response Mean        & 31.8\% & 56.4\% \\
Worst-10\% Mean           & 34.3\% & 60.8\% \\
\textbf{Full-Response Min}& \textbf{35.2\%} & \textbf{62.4\%} \\
Oracle                   & 56.4\% & 100\% \\
\bottomrule
\end{tabular}
\end{table}

\subsection{Self-Correction with Token-Level Hints}
\label{appendix:self_correction}

We investigate whether token-level error localization from \textsc{TokenHD} can assist a language model in correcting its own mistakes. For each incorrect sample, a correction model (we use GPT-4o-mini to perform the self-correction since it is the primary policy model used in our experiments) is provided with the original problem, the previous incorrect solution, and a condition-specific hint. We run up to three correction iterations per sample: if the correction model produces the correct answer on iteration $k$, the sample is counted as successfully corrected at iteration $k$; otherwise, the corrected solution from iteration $k$ is fed back as the new ``previous solution'' for iteration $k+1$. We compare four conditions that differ in the type of hint provided.

\textbf{Baseline}: the correction model is informed that its previous answer is incorrect and asked to retry, with no localization information provided. \textbf{TokenHD}: suspected error regions are highlighted directly in the solution text using inline markers (\texttt{<<<...>>>}), generated by \textsc{TokenHD}-8B (prompt design shown in Figure~\ref{fig:self_correction_prompt}). The hints direct the correction model's attention to the most suspect token spans. \textbf{Step}: errors are marked at the paragraph level by aggregating token-level scores to the nearest paragraph boundary, simulating the coarser granularity of a PRM-style hint. \textbf{Oracle}: ground-truth error spans are directly provided as hints, representing the theoretical performance upper bound.

As shown in Table~\ref{tab:self_correction}, the baseline correction rate is 16.9\%, reflecting the difficulty of self-correction without any localization cues. Both the TokenHD condition (19.9\%) and the step condition (19.1\%) significantly outperform the baseline, demonstrating that even approximate error localization enables more targeted revision. Crucially, \textsc{TokenHD} reaches 98\% of the oracle upper bound (19.9\% vs.\ 20.3\%), confirming that fine-grained token-level localization translates directly into effective correction guidance. The higher first-iteration success rate for TokenHD (11.4\%) compared to the step condition (8.9\%) further shows that finer-grained localization enables the correction model to identify and fix errors more immediately, reducing the number of revision attempts required. Together, these results confirm that \textsc{TokenHD} provides a practical and efficient mechanism for targeted self-correction, approaching oracle-level performance while requiring no ground-truth information.

\begin{table}[h]
\centering
\caption{Self-correction results on incorrect Math-500 samples using GPT-4o-mini as the correction model, run for up to three iterations. Correction Rate: fraction of samples successfully corrected. 1st-Iter Rate: fraction corrected on the first iteration. \% of Oracle: correction rate as a fraction of the oracle upper bound (20.3\%).}
\label{tab:self_correction}
\begin{tabular}{lccc}
\toprule
\textbf{Condition} & \textbf{Correction Rate} & \textbf{1st-Iter Rate} & \textbf{\% of Oracle} \\
\midrule
Baseline & 16.9\% & 8.5\%  & 83.3\% \\
Step     & 19.1\% & 8.9\%  & 94.1\% \\
TokenHD  & 19.9\% & 11.4\% & \textbf{98.0\%} \\
Oracle   & 20.3\% & 14.4\% & 100\% \\
\bottomrule
\end{tabular}
\end{table}

\begin{figure}[ht]

\begin{AIBox}{Prompt for Self-Correction with Token-Level Hints}

\parbox[t]{\textwidth}{

\scriptsize \begin{alltt}
You are given a math problem and a previous solution attempt that was incorrect.
Suspected error regions in the previous solution are marked with triple angle
  brackets:
<<<error region>>>.

Your task is to carefully review the previous solution, paying special attention
to the marked regions, identify the mistakes, and produce a corrected solution.


**Instructions**:

- Re-examine the flagged regions carefully; they are likely where the error occurred.

- Provide a complete corrected solution, not just the corrected portions.

- Ensure your final answer is placed inside \textbackslash{}boxed\{\}.


**Input Format**:

Problem: [PROBLEM\_STATEMENT]

Previous Solution (with error regions marked):
[PREVIOUS\_SOLUTION\_WITH\_<<<...>>>\_MARKERS]


**Output Format**:

Corrected Solution: [YOUR\_CORRECTED\_SOLUTION]


Please analyze and correct the following:


Problem: \{problem\}


Previous Solution:
\{previous\_solution\_with\_hints\}
\end{alltt}}

\end{AIBox}
\caption{Prompt used for self-correction with token-level hints. Suspected error regions identified by \textsc{TokenHD}-8B are highlighted inline using \texttt{<<<...>>>} markers, directing the correction model's attention to the most suspect token spans.}
\label{fig:self_correction_prompt}
\end{figure}

\section{Robustness of the Evaluation Protocol}
\label{appendix:eval_robustness}

Our evaluation relies on two design choices that could each affect the validity of reported results: the ensemble weights that aggregate critic annotations into training labels, and the binarization thresholds used for evaluation. We conduct two validation experiments to confirm that these choices are stable.

\subsection{Ensemble Weight Stability Across Data Subsets}
\label{appendix:weight_stability}

Our adaptive ensemble assigns learned weights to each critic model by minimizing Eq.~\ref{eq:ensemble_loss} on a held-out validation set $\mathcal{D}_{\mathrm{val}}$. To verify that these weights are stable and do not depend on the specific data used for optimization, we repeat the weight optimization on two independent held-out subsets from our training data: Math-Train (drawn from the Math dataset) and AceReason (drawn from AceReason-Math). Both subsets are mutually disjoint from each other and from the samples used for model training. We optimize the weights separately on each subset using the same procedure described in Section~\ref{subsec:token_ensemble}. To quantify variability across data samples within each subset, we apply bootstrap resampling: for each subset, we draw 1{,}000 bootstrap resamples by sampling the same number of individual (query, response, critic annotation) tuples with replacement as in the original subset, run the weight optimization on each resample, and compute 95\% confidence intervals from the 2.5th and 97.5th percentiles of the resulting weight distributions.

The weight ordering is consistent across both subsets. o4-mini receives the highest weight (${\approx}0.60$; 95\% CI: [0.52, 0.65] on Math-Train, [0.53, 0.64] on AceReason), followed by GPT-4.1 (${\approx}0.30$) and QwQ-32B (${\approx}0.09$). The point estimates differ by less than 2\% across the two subsets for all critics. These results show that the learned weights are not sensitive to the choice of optimization data, and that the assigned weights reflect genuine capability differences among the critics rather than subset-specific noise.

\subsection{Threshold Sensitivity}
\label{appendix:threshold_sensitivity}

We use fixed binarization thresholds $\beta_I = 0.5$ for ground-truth labels and $\beta_{\widehat{I}} = 0.5$ for predicted scores throughout all experiments. To assess how sensitive the reported results are to this choice, we evaluate \textsc{TokenHD}-8B on all five benchmarks under perturbed threshold settings. Specifically, we vary each threshold one at a time over three values $\{0.45, 0.50, 0.55\}$ while holding the other fixed at its default of 0.5. This yields four non-default configurations in total: $(\beta_I, \beta_{\widehat{I}}) \in \{(0.45, 0.5),\,(0.55, 0.5),\,(0.5, 0.45),\,(0.5, 0.55)\}$. For each configuration, we compute the absolute deviation in $S_{\textrm{incor}}$ relative to the default setting $(\beta_I{=}0.5,\,\beta_{\widehat{I}}{=}0.5)$, and report the maximum deviation across all four configurations for each benchmark in Table~\ref{tab:threshold_sensitivity}.

\begin{table}[h]
\centering
\caption{Maximum absolute $S_{\textrm{incor}}$ deviation (percentage points) from the default threshold pair $(\beta_I{=}0.5,\,\beta_{\widehat{I}}{=}0.5)$ when each threshold is independently varied by $\pm0.05$. Default $S_{\textrm{incor}}$ and the observed range are shown for reference.}
\label{tab:threshold_sensitivity}
\setlength{\tabcolsep}{5pt}
\begin{tabular}{lcccc}
\toprule
\textbf{Benchmark} & \textbf{Default (\%)} & \textbf{Range (\%)} & \textbf{Max $|\Delta|$ (pts)} \\
\midrule
Math-500    & 60.73 & [59.65, 61.26] & 1.08 \\
Olym-Math   & 65.42 & [64.78, 65.42] & 0.64 \\
GPQA        & 63.47 & [61.81, 64.63] & 1.66 \\
Olym-Phy    & 63.68 & [63.51, 64.46] & 0.78 \\
FinQA       & 61.34 & [57.97, 63.47] & 3.37 \\
\midrule
\textbf{Mean} & $-$ & $-$ & \textbf{1.51} \\
\bottomrule
\end{tabular}
\end{table}

For the four mathematical and scientific reasoning benchmarks (Math-500, Olym-Math, GPQA, Olym-Phy), the maximum deviation is at most 1.66 percentage points. FinQA, the specialized financial QA benchmark, shows a larger maximum deviation of 3.37 percentage points; the mean across all five benchmarks is 1.51 percentage points. These results confirm that our evaluation is not sensitive to the exact threshold values, and that the reported performance differences remain stable under reasonable threshold perturbations.

\section{Detection Performance Across Response Lengths}
\label{appendix:length_analysis}

We examine whether detection performance varies with response length. We partition the incorrect evaluation samples into three bins by absolute token count ($<$500, 500--1000, $>$1000) and report the average $S_{\textrm{incor}}$ across all seven benchmarks for both \textsc{TokenHD}-1.7B and \textsc{TokenHD}-8B.

\begin{figure}[h]
    \centering
    \includegraphics[width=0.58\textwidth]{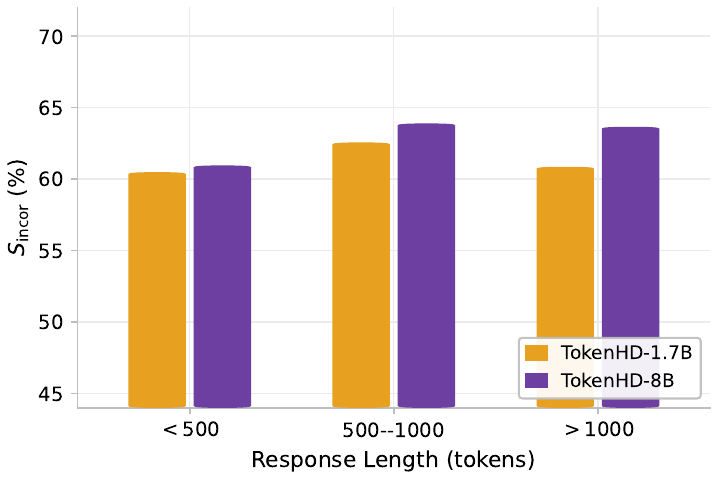}
    \caption{Average $S_{\textrm{incor}}$ across seven benchmarks for \textsc{TokenHD}-1.7B and \textsc{TokenHD}-8B, grouped by response length. Samples are partitioned by absolute token count into three bins: $<$500, 500--1000, and $>$1000.}
    \label{fig:length_analysis}
\end{figure}

As shown in Figure~\ref{fig:length_analysis}, both models achieve their highest $S_{\textrm{incor}}$ in the 500--1000 token bin (\textsc{TokenHD}-1.7B: 62.6, \textsc{TokenHD}-8B: 63.9). Both models show modest declines in the $<$500 token bin (\textsc{TokenHD}-1.7B: 60.5, \textsc{TokenHD}-8B: 61.0). In the $>$1000 token bin, \textsc{TokenHD}-1.7B shows a similar decline (60.9), while \textsc{TokenHD}-8B remains close to its peak (63.7). \textsc{TokenHD}-8B consistently outperforms \textsc{TokenHD}-1.7B across all bins, in line with the scaling trend observed in the main results.

\newpage

\end{document}